\documentclass{article}

\usepackage[preprint]{corl_2026} 

\usepackage{amsfonts}
\usepackage{amsmath}
\usepackage{booktabs}
\usepackage{caption}
\usepackage{enumitem}
\usepackage{graphicx}
\usepackage{multirow}
\usepackage{multicol}
\usepackage{xcolor}
\usepackage{float}

\definecolor{citecolor}{HTML}{0071BC}
\hypersetup{colorlinks,linkcolor={black},citecolor={citecolor}}  

\title{Current as Touch: Proprioceptive Contact Feedback for Compliant Dexterous Manipulation}

\author{
  Chenyang Ma\thanks{Equal contribution.}, 
  Yunchao Yao\footnotemark[1], 
  Zhenyu Wei\footnotemark[1], 
  Ruogu Li, 
  Daniel Szafir\thanks{Equal advising.}, 
  Mingyu Ding\footnotemark[2]
  \\
  University of North Carolina at Chapel Hill
  \\
  \texttt{
    \{mach, yunchaoy, wzhenyu, ruogu, dszafir, md\}@cs.unc.edu
  }
}

\begin{document}
\maketitle
\hypersetup{linkcolor={red}}

\begin{center}
    \vspace{-12pt}
    \includegraphics[width=\linewidth]{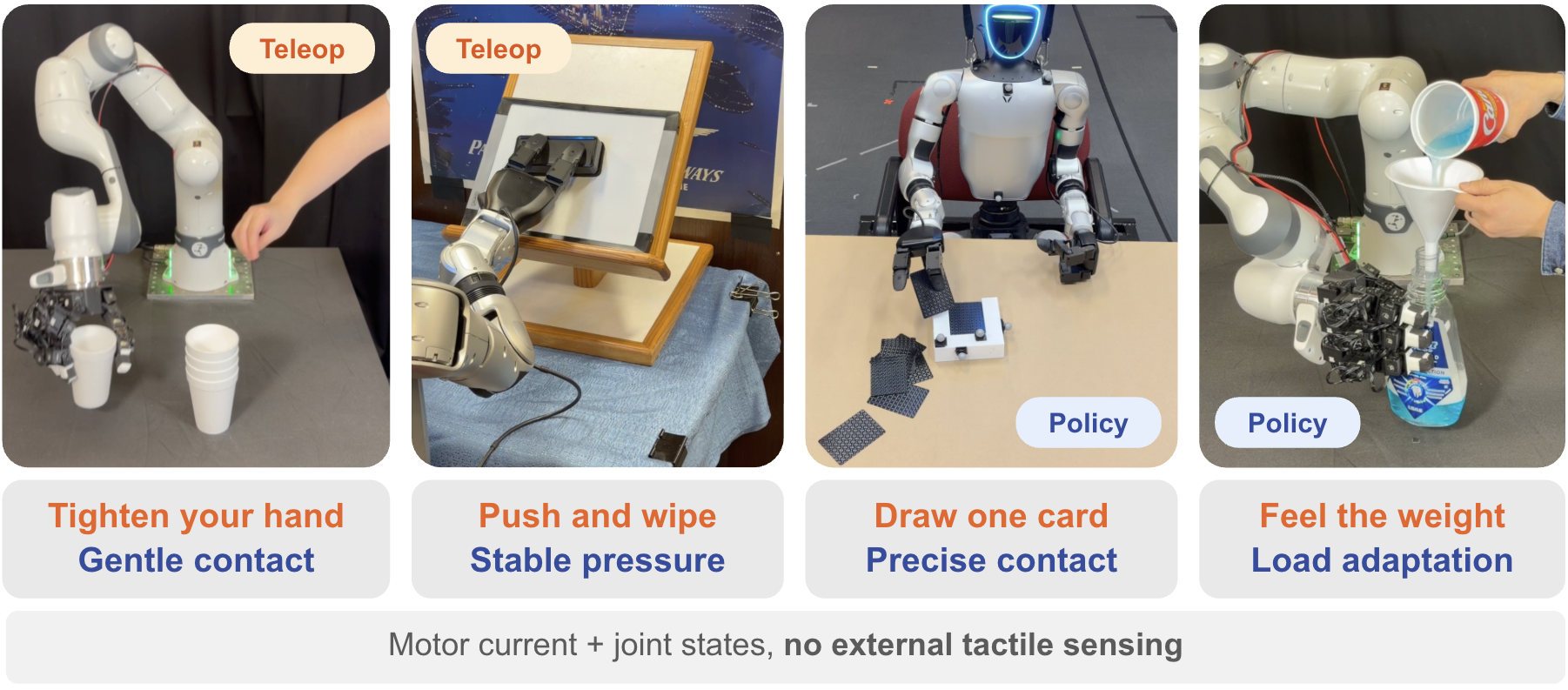}
    \vspace{-16pt}
    \captionof{figure}{\textbf{Current as Touch for contact-rich dexterous manipulation.}
    Our proprioception-driven compliance framework uses motor current and joint states as tactile-like contact cues, enabling compliant manipulation without external tactile sensors.
    Across teleoperation and policy-learning settings, the same compliance-reference prediction interface supports fragile object handling, sustained surface contact, thin-object retrieval, and dynamic load adaptation.}
    \vspace{6pt}
    \label{fig:teaser}
\end{center}


\begin{abstract}
Compliance is essential for dexterous manipulation, yet existing solutions often rely on external tactile or force sensors that are costly, fragile, and difficult to deploy on low-cost robot hands. We propose a proprioception-driven framework that learns contact-aware compliance cues from motor current and joint states. Since motor current is closely related to actuator torque, it provides an intrinsic signal for perceiving contact force, object resistance, and grasp stability without additional sensing hardware.
Rather than estimating external wrenches or commanding torque, our method predicts a compliance reference position: an ideal joint-position target for a standard PD controller whose induced position error generates appropriate grasping force.
This position-based formulation is compatible with mainstream teleoperation and policy-learning pipelines, while enabling the robot to adapt interaction forces from real-time proprioceptive feedback.
Thus, motor current serves not only as a force proxy but also as a learnable proprioceptive contact signal for compliance reference prediction.
Experiments on multiple dexterous hands and contact-rich tasks, including fragile object handling, sustained surface contact, thin-object retrieval, and dynamic load adaptation, show stable compliant grasping, safer and more efficient teleoperation, and improved downstream policy learning without external tactile or force sensors. More details and demos can be found at our webpage: \url{https://cat.chenyangma.com/}.
\end{abstract}



\section{Introduction}

Dexterous manipulation in the real world requires more than accurate position tracking. When a robot hand interacts with fragile, deformable, or dynamically changing objects, it must continuously regulate contact forces while maintaining stable grasping~\cite{okamura2000overview, wei2024dro}. As illustrated in Fig.~\ref{fig:teaser}, representative contact-rich scenarios include stacking paper cups, which requires gentle contact to avoid deformation; wiping a board, which requires sustained pressure; retrieving a single card, which requires precise interaction with thin objects; and pouring water, which requires adapting to changing load. These tasks are difficult for purely rigid position control, where an inaccurate target position can easily generate excessive force or cause slip.

A common way to achieve compliance is to equip robot hands with tactile sensors or force/torque sensors~\cite{yuan2017gelsight, lambeta2020digit, bhirangi2021reskin, bhirangi2025anyskin, choi2025coinft}. These sensors provide direct measurements of contact and enable force-aware control. However, they also introduce additional cost, fragility, calibration effort, and integration complexity. This limits their deployment on many low-cost dexterous hands, which are usually driven by joint-level position commands and low-level PD controllers. As a result, there remains a gap between the compliant interaction required by contact-rich manipulation and the hardware interfaces available in widely used robot hands.

\begin{figure}[htbp]
    \centering
    \includegraphics[width=1.0\linewidth]{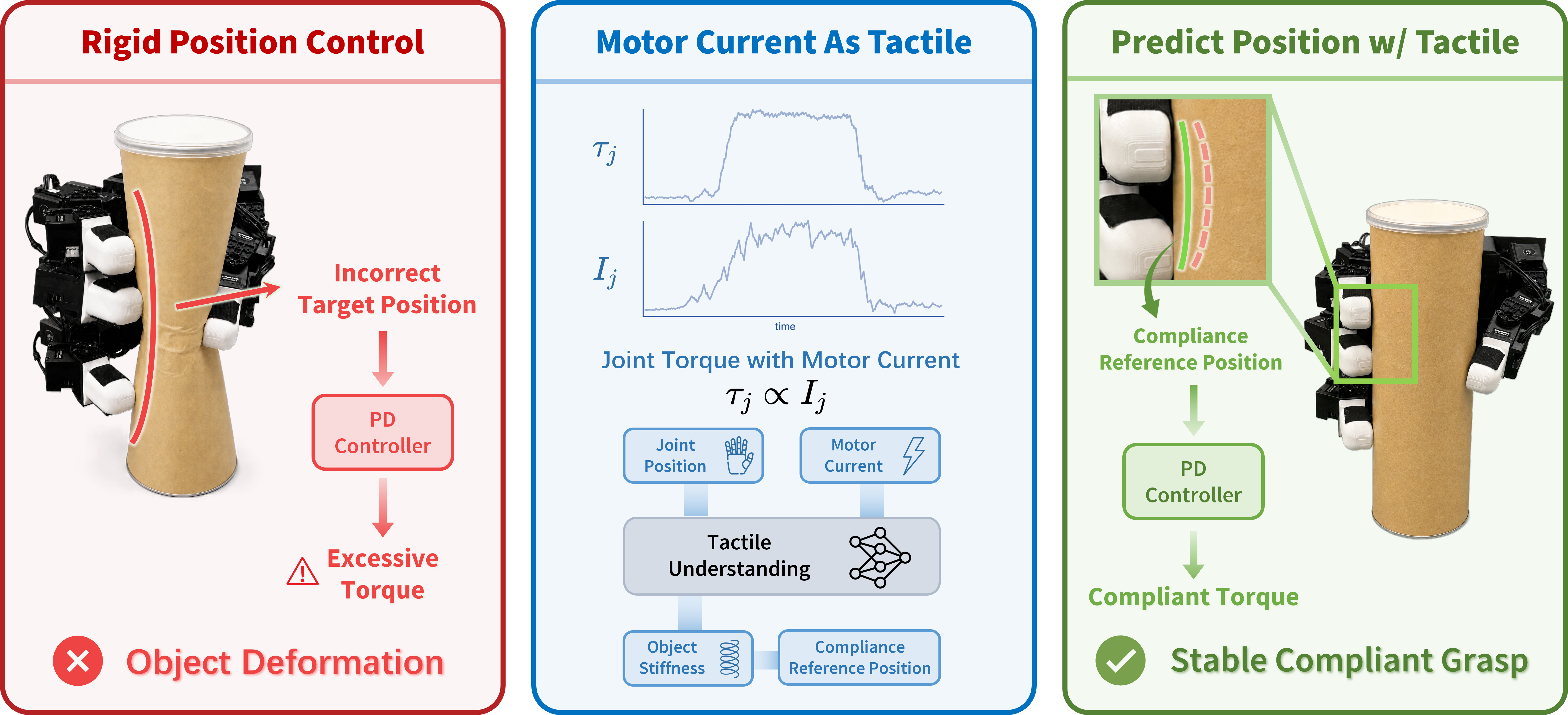}
    \caption{
    \textbf{Motivation for Current as Touch.}
    Rigid position control may generate excessive contact forces and damage fragile objects when target positions are inaccurate.
    Motor current and joint states offer built-in tactile-like proprioceptive feedback about contact and object resistance.
    We use this feedback to predict a compliance reference position (CRP), allowing standard PD position control to produce compliant torque without external tactile/force sensors or direct torque commands.
    }
    \label{fig:motivation}
\end{figure}

Fig.~\ref{fig:motivation} summarizes our motivating observation and design goal. In a standard position-controlled hand, an inaccurate target position can penetrate a fragile or deformable object, and the resulting PD tracking error may generate excessive torque. At the same time, in motor-driven dexterous hands, contact forces are generated through actuator torques, which are closely related to motor current. Motor current, together with joint states, therefore provides a built-in proprioceptive signal that reflects 
useful contact cues, such as contact resistance, force, grasp loading, and changes in object interaction.
This motivates us to ask whether tactile-like contact feedback for compliance control can be obtained from signals intrinsic to the hand, without relying on external tactile or force sensors.

\begin{figure}[t]
    \centering
    \includegraphics[width=1.0\linewidth]{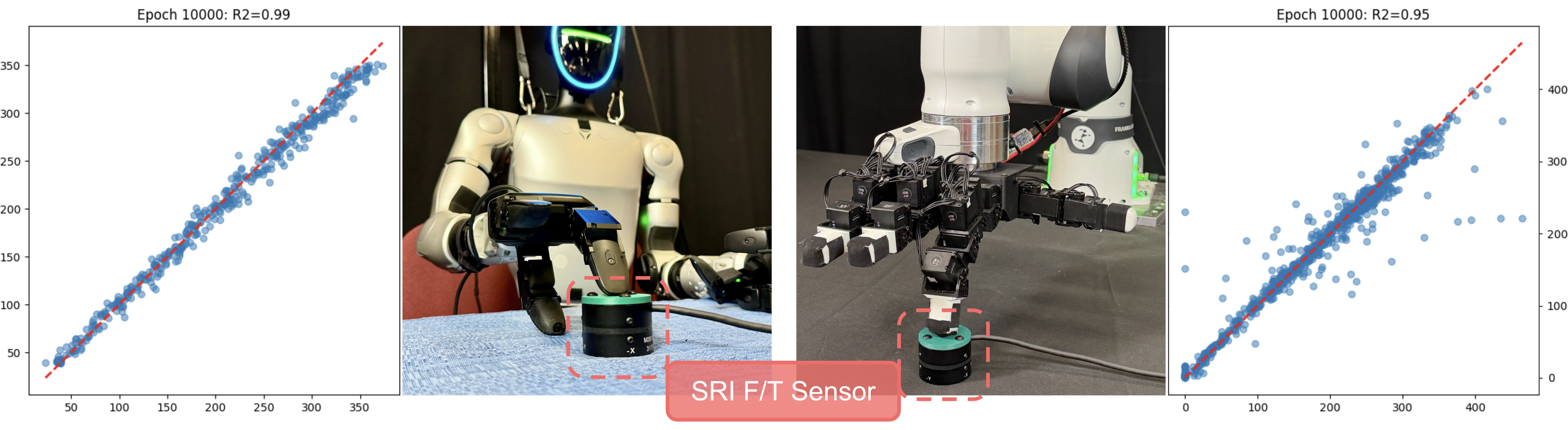}
    \vspace{-16pt}
    \caption{
    \textbf{Motivating measurement of motor current as proprioceptive contact feedback.}
    On Unitree Dex3~\cite{unitreeDex3} and LEAP Hand~\cite{shaw2023leaphand}, motor current and joint states vary consistently with contact force measured by an external force/torque sensor. This observation motivates using intrinsic actuator signals as contact feedback without treating them as explicit force estimates.
    }
    \vspace{-6pt}
    \label{fig:scale_experiment}
\end{figure}

As a motivating empirical observation, Fig.~\ref{fig:scale_experiment} illustrates that motor current and joint states vary consistently with measured contact force across different dexterous hands. A simple current-and- position regressor predicts measured normal force with RMSEs of 10.09g on Dex3 and 17.75g on LEAP Hand, with $R^2$ values of 0.99 and 0.95, respectively. This suggests that intrinsic motor signals contain useful contact information, motivating their use as proprioceptive feedback rather than explicit force estimates.

However, the goal of compliant manipulation is not merely to estimate contact force. Most teleoperation systems and learned policies for dexterous hands do not command torque directly; they output target joint positions that are tracked by low-level PD controllers. This creates a mismatch between tactile-like proprioceptive feedback, which reveals physical interaction, and the action space used by existing robot learning pipelines. To make such feedback practically useful, it must be converted into a position reference that can induce appropriate compliant torques through standard PD control.

In this work, we propose a proprioception-driven compliance framework that predicts a compliance reference position from motor current and joint states rather than direct torque commands.
This reference is the ideal joint-position target sent to the PD controller: when tracked by the controller, the resulting position error generates the appropriate grasping force for stable contact without crushing the object. In this sense, compliance is represented as an adaptive position target: the robot still acts through the standard position-control interface, but the commanded target is informed by torque-aware proprioceptive feedback. 
This formulation allows current-based contact feedback to resolve contact ambiguity in position-controlled manipulation: for similar hand poses, different current responses can lead to different compliance references and user intents. The resulting grasp representation can be a general interface compatible with both teleoperation and policy learning.

We evaluate our framework on multiple dexterous hands and diverse contact-rich tasks, spanning fragile object handling, sustained surface contact, thin-object retrieval, and dynamic load adaptation. The results show that proprioceptive current feedback enables stable compliant grasping under varying contact conditions, improves teleoperation safety and efficiency, and benefits downstream policy learning across different hands and tasks.

In summary, the main contributions of this work are threefold:
\begin{itemize}
    \item We learn torque-aware proprioceptive contact feedback from motor current and joint states, enabling tactile-free compliant manipulation without external tactile or force sensors.

    \item We formulate compliant grasping as compliance-reference prediction, which converts contact-aware proprioceptive feedback into position references that 
    induce compliant torques through standard PD controllers, while remaining compatible with position-based teleoperation and policy learning.

    \item We validate this proprioceptive compliance-reference interface across multiple dexterous hands and contact-rich tasks, showing consistent gains in teleoperation safety, efficiency, and downstream policy robustness without external tactile or force sensors.

\end{itemize}

\section{Related Work}

\textbf{Tactile and Force-Torque Sensing for Contact-Rich Manipulation.}
Tactile and force sensing provide contact information for dexterous manipulation~\cite{wei2026one,liang2025dexhanddiff,bao2023dexart,wang2023dexgraspnet,xu2023unidexgrasp,wan2023unidexgrasppp,zhang2025dexgraspnet2,zhang2026unidex,zhao2025dexh2r}, contact-rich manipulation~\cite{huang2026dexcompose,li2026coordex}, where vision and position tracking often miss local geometry, slip, shear, and interaction forces.
Prior work has developed high-resolution tactile fingertips such as GelSight~\cite{yuan2017gelsight}, DIGIT~\cite{lambeta2020digit}, soft rounded tactile sensors~\cite{romero2020soft}, DTact~\cite{lin2022dtact}, and 9DTact~\cite{lin20239dtact}, as well as scalable tactile skins and arrays such as ReSkin~\cite{bhirangi2021reskin}, AnySkin~\cite{bhirangi2025anyskin}, and flexible tactile arrays~\cite{zhao2022large}.
Recent learning systems incorporate tactile or visuo-tactile observations into manipulation policies, including 3D-ViTac~\cite{huang20243d}, multi-modal tactile diffusion policies~\cite{zhao2025polytouch}, and bimanual or reactive policies using tactile feedback~\cite{gu2025tactilealona, xue2025reactive}.
Force/torque sensing similarly provides direct wrench measurements, with compact or finger-mounted systems such as CoinFT~\cite{choi2025coinft} and UMI-FT~\cite{choi2026wild} enabling force-aware manipulation learning.
While effective, these approaches require additional tactile or force-torque hardware, calibration, and integration. In contrast, we study whether contact-aware signals can be recovered from motor current and joint states already available inside dexterous hands, avoiding external tactile or force/torque sensors.

\textbf{Compliance Control with Proprioceptive Signals.}
Compliant manipulation has been extensively studied through hybrid position/force control, impedance control, and variable impedance control, which regulate robot motion based on physical interaction with the environment~\cite{raibert1981hybrid, hogan1985impedance, anderson1988hybrid, siciliano1999robot, abu2020variable}. 
Although these methods enable stable and safe contact, they often require force/torque sensing, accurate dynamics, or torque-level control. Sensorless force estimation and current-based interaction control reduce this reliance by inferring external forces or joint torques from motor currents, joint states, disturbance observers, or actuator models~\cite{wahrburg2017motor, liu2021sensorless, han2022sensorless}.
Recent work has further explored compliant grasp planning and learned adaptive compliance for contact-rich manipulation~\cite{chen2024springgrasp, hou2025adaptive}, and has begun to exploit motor-current signals for dexterous manipulation and compliance. For example, Zhao et al.~\cite{zhao2026closing} calibrate motor current into joint torque and combine it with dense tactile feedback in a sim-to-real RL framework for force-based dexterous grasping and manipulation. While this suggests that actuator-level signals capture useful physical interaction information, the method still relies on dense tactile observations, tactile simulation, and task-specific reinforcement learning. More closely related to our work, Minimalist Compliance Control~\cite{shi2026minimalist} estimates external wrenches from motor current or voltage signals through motor models and robot Jacobians, and then updates position references with a task-space admittance controller. Although it demonstrates that actuator signals can substitute for external F/T sensors in compliance control, it still follows an explicit wrench-estimation and model-based admittance-control formulation.

In contrast, our method learns compliance reference positions directly from motor current and joint states, avoiding explicit wrench recovery while matching the joint-position command interface used by dexterous teleoperation and policy learning.

\section{Methodology}

Motivated by the empirical trend in Fig.~\ref{fig:scale_experiment}, we use motor current, together with joint position, as proprioceptive contact feedback. Our goal is not to explicitly recover contact force or external wrench. Most dexterous hands expose a position-control interface, where a low-level PD controller tracks commanded joint references. Therefore, we formulate compliance as a reference-position prediction problem: given proprioceptive contact feedback, the model predicts a joint-position reference whose induced PD torque produces appropriate contact force.

\subsection{Compliance Reference Position}

Let $q_t$ denote the measured joint position of the hand, $\dot q_t$ the joint velocity, and $q^{c}_{\mathrm{ref},t}$ the compliance reference position (CRP) predicted by our model. The low-level hand controller tracks this reference through a standard PD law:
\begin{equation}
    \tau_t = K_p \left(q^{c}_{\mathrm{ref},t} - q_t\right) - K_d \dot q_t ,
\end{equation}
where $\tau_t$ is the low-level controller torque, and $K_p$ and $K_d$ are fixed PD gains.

This formulation is particularly suitable for low-cost dexterous hands. If the reference is too far inside a rigid or fragile object, the PD controller produces excessive torque and may crush the object. If the reference is too close to the current position, the hand may fail to generate sufficient normal force and slip. The desired CRP therefore depends on the current hand state, contact condition, object response, and task intent. Our method learns this reference directly from data, using motor current as proprioceptive feedback about contact resistance.

\begin{figure}[t]
    \centering
    \includegraphics[width=0.92\linewidth]{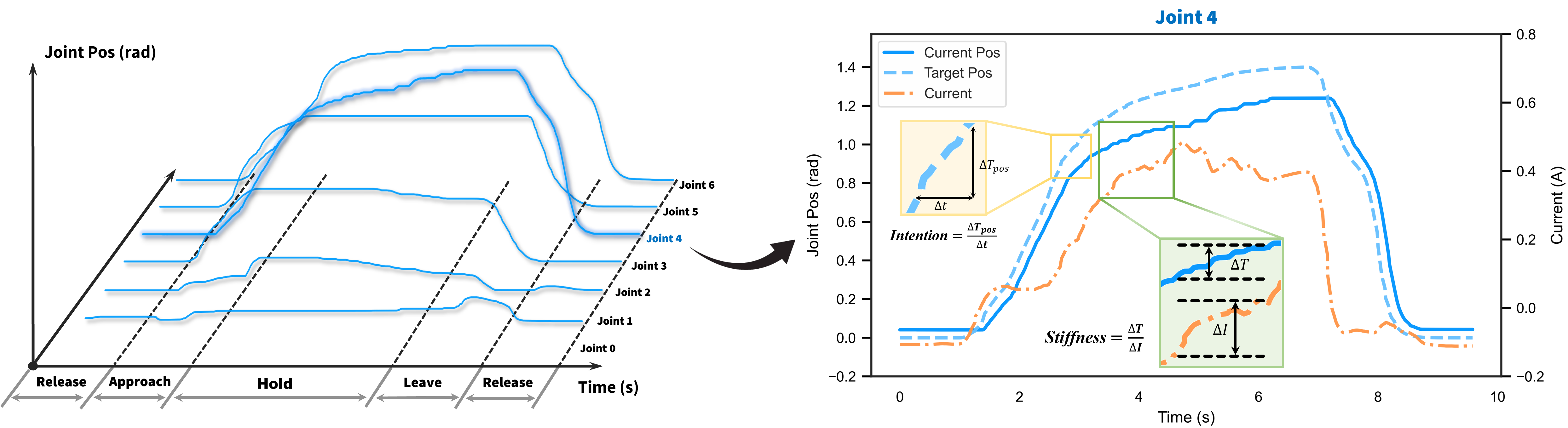}
    \vspace{-3pt}
    \caption{
    \textbf{Teleoperation data manifold for learning CRPs.}
    Demonstrations couple user intent with contact responses: command slopes encode grasp/release intent, while current changes relative to joint motion, e.g., $\Delta I / \Delta q$, provide stiffness-related contact cues. The model learns CRPs from this human-corrected manifold.
    }
    \vspace{-12pt}
    \label{fig:data_manifold}
\end{figure}

\subsection{Learning CRPs from Human-in-the-Loop Demonstrations}

A key question is how we supervise $q^{c}_{\mathrm{ref}}$. We use human teleoperation demonstrations as a human-in-the-loop closed-loop correction. During data collection, the operator does not command force directly. Instead, the operator observes the task outcome, such as object deformation, slip, or stable grasping, and continuously adjusts the target joint position $q_{\mathrm{cmd}}$. As a result, the recorded command is not an arbitrary open-loop position target. It is a noisy but valid reference that has been corrected through visual feedback and whose induced PD torque achieves successful compliant interaction.

We therefore use the demonstrated command trajectory as supervision for the CRP:
\begin{equation}
    q^{c,*}_{\mathrm{ref},t} = q_{\mathrm{cmd},t}.
\end{equation}
These labels are not analytical optima; rather, they are noisy but task-valid CRP targets produced by human-in-the-loop closed-loop correction. Importantly, during inference, the model should not simply copy user command. Directly feeding $q_{\mathrm{cmd},t}$ as input would leak the target into the observation and encourage shortcut learning. Instead, we represent user intent by the command velocity:
\begin{equation}
    v_{\mathrm{intent},t} = q_{\mathrm{cmd},t} - q_{\mathrm{cmd},t-1}.
\end{equation}
This intent velocity captures whether the user is trying to grasp, release, or maintain contact, as well as how aggressively the user wants the hand to move. The model must combine this intent with the current hand state and motor-current feedback to predict the appropriate CRP.

Fig.~\ref{fig:data_manifold} illustrates why such supervision contains learnable compliance signals. During free-space motion, position changes induce only small current variation, reflecting internal friction and actuator dynamics. When the hand contacts an object, small position changes can cause large current changes. The local relationship between $\Delta q$ and $\Delta I$ provides contact stiffness information. Meanwhile, the slope of the user command trajectory provides grasping or releasing intent. Thus, the demonstration manifold contains both intent and contact-response information, allowing the model to learn how to map proprioceptive interaction signals to compliant reference positions.

\begin{figure}[t]
    \centering
    \includegraphics[width=0.95\linewidth]{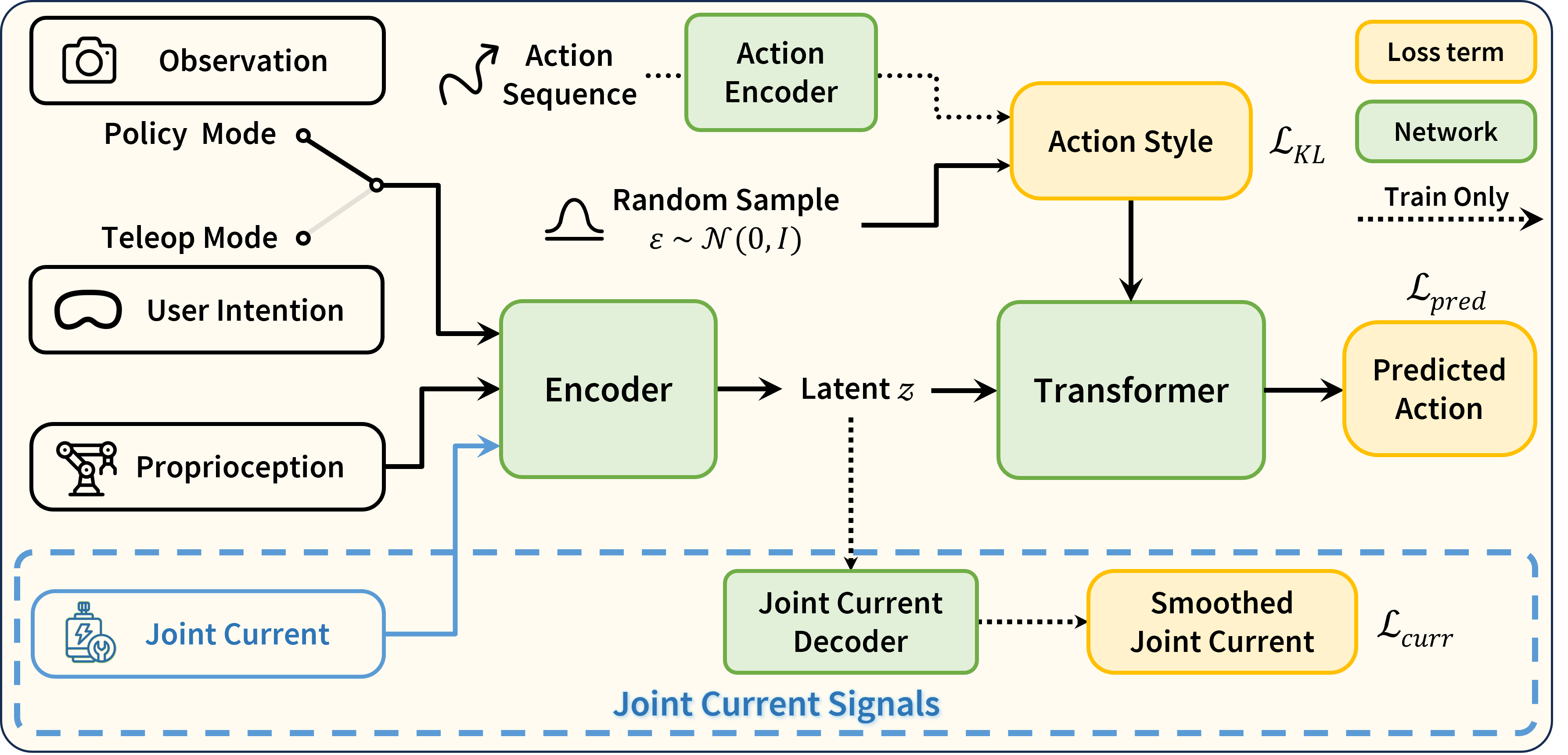}
    \caption{
    \textbf{Current-conditioned prediction pipeline.}
    The model predicts CRPs from proprioceptive history. In teleoperation, inputs include robot state, raw motor current, and user intent velocity; in policy mode, user intent is replaced by task-level object or goal pose. 
    }
    \label{fig:method_pipeline}
\end{figure}

\subsection{Current-Conditioned Teleoperation and Policy Learning}

Our framework supports two execution modes, as shown in Fig.~\ref{fig:method_pipeline}. In teleoperation mode, the model assists a human operator by converting intent into a contact-aware CRP. Given an observation history of length $T$, the model receives
\begin{equation}
    o^{\mathrm{teleop}}_t =
    \left\{
    q_{t-T+1:t},\;
    I_{t-T+1:t},\;
    v_{\mathrm{intent},t-T+1:t}
    \right\},
\end{equation}
where $I_t$ is the raw motor current. The model predicts the next compliance reference position:
\begin{equation}
    \hat q^{c}_{\mathrm{ref},t} = f_\theta(o^{\mathrm{teleop}}_t),
\end{equation}
where $f_\theta$ denotes the learned CRP predictor parameterized by $\theta$, $q_t$ specifies the hand configuration, $I_t$ provides contact-dependent feedback, and $v_{\mathrm{intent},t}$ specifies the direction and speed of the user's intended motion. The predicted CRP then drives the hand through the standard PD controller.

In policy-learning mode, there is no online user command. The model instead conditions on task-level perception, such as the object or goal pose $g_t \in SE(3)$, together with proprioceptive feedback:
\begin{equation}
    o^{\mathrm{policy}}_t =
    \left\{
    q_{t-T+1:t},\;
    I_{t-T+1:t},\;
    g_t
    \right\}.
\end{equation}
The demonstrated action is again used as the CRP supervision. In this setting, perception provides the task state, while motor current provides the contact state. This allows the policy to produce different reference actions for the same visual state depending on whether the hand has established contact, is slipping, is overloaded, or has reached a stable grasp.

Both modes are trained with a reference prediction loss:
\begin{equation}
    \mathcal{L}_{\mathrm{ref}}
    =
    \left\|
    \hat q^{c}_{\mathrm{ref},t}
    -
    q^{c,*}_{\mathrm{ref},t}
    \right\|_2^2 .
\end{equation}
Our implementation follows ACT-style sequence modeling. As shown in Fig.~\ref{fig:method_pipeline}, we retain the action encoder and KL regularization during training to capture variability in reference actions, while focusing on the current-conditioned observation encoder and auxiliary current supervision.

Although our network follows the sequence-modeling spirit of ACT, we mainly use temporal context on the observation side to capture recent current and motion trends. During real-robot execution, we apply a short exponential moving average over the most recent two predictions to reduce hardware jitter while preserving responsiveness to current changes.

\subsection{Raw-Current Encoding with Smoothed-Current Auxiliary Supervision}

Motor current is informative but noisy. Raw current often contains high-frequency PWM noise, communication spikes, and actuator disturbances. A straightforward solution would be to low-pass filter the current before feeding it into the policy, but filtering introduces phase delay, which weakens the response to sudden contact changes. We therefore feed raw, zero-delay current into the model.

Let $\bar I_t$ denote an offline-smoothed current used only for supervision.
Given the latent representation $z_t$ from the observation encoder, an auxiliary decoder
$h_\phi$ predicts the current and is trained with:
\begin{equation}
    \hat{\bar I}_t = h_\phi(z_t), \qquad
    \mathcal{L}_{\mathrm{cur}}
    =
    \left\|
        \hat{\bar I}_t - \bar I_t
    \right\|_2^2.
\end{equation}

\begin{table}[t]
    \centering

    \begin{minipage}[t]{0.58\linewidth}
        \centering
        \setlength{\tabcolsep}{4pt}
        \renewcommand{\arraystretch}{0.9}
        \small
        \caption{Teleop foam-cup stacking results.}
        \vspace{3pt}
        \resizebox{\linewidth}{!}{%
            \begin{tabular}{lcccc}
                \toprule
                Operator & Method & Time (s) $\downarrow$ & Deformed (\%) $\downarrow$ & Grasp Failure (\%) $\downarrow$ \\
                \midrule
                \multirow{3}{*}{Novice}
                    & Retargeting & 21.9 & 43.3\% & 13.3\% \\
                    & w/o Current & 31.1 & 15.0\% & 76.7\% \\
                    & w/ Current & \textbf{16.8} & \textbf{0.0\%} & \textbf{6.7\%} \\
                \midrule
                \multirow{3}{*}{Skilled}
                    & Retargeting & 20.7 & 25.0\% & 15.0\% \\
                    & w/o Current & 35.3 & 16.7\% & 71.7\% \\
                    & w/ Current & \textbf{16.1} & \textbf{0.0\%} & \textbf{3.3\%} \\
                \bottomrule
            \end{tabular}%
        }
        \label{tab:foam_cup}
    \end{minipage}
    \hfill
    \begin{minipage}[t]{0.4\linewidth}
        \centering
        \setlength{\tabcolsep}{4pt}
        \renewcommand{\arraystretch}{0.9}
        \small
        \caption{Teleop board-wiping results.}
        \vspace{3pt}
        \resizebox{\linewidth}{!}{%
            \begin{tabular}{lccc}
                \toprule
                Operator & Method & Success (\%) $\uparrow$ & Time (s) $\downarrow$ \\
                \midrule
                \multirow{3}{*}{Novice}
                    & Retargeting & 40.0 & 9.3 \\
                    & w/o Current & 46.7 & 9.3 \\
                    & w/ Current & \textbf{100.0} & \textbf{6.9} \\
                \midrule
                \multirow{3}{*}{Skilled}
                    & Retargeting & \textbf{100.0} & 7.1 \\
                    & w/o Current & 60.0 & 8.3 \\
                    & w/ Current & \textbf{100.0} & \textbf{6.5} \\
                \bottomrule
            \end{tabular}%
        }
        \label{tab:whiteboard_wiping_results}
    \end{minipage}

    \vspace{-12pt}
\end{table}
\begin{table}[t]
    \centering
    \begin{minipage}[t]{0.48\linewidth}
        \centering
        \setlength{\tabcolsep}{3pt}
        \renewcommand{\arraystretch}{0.9}
        \scriptsize
        \caption{Dynamic bottle-holding results.}
        \vspace{3pt}
        \resizebox{\linewidth}{!}{%
            \begin{tabular}{ccccc}
                \toprule
                \textbf{Water} & \textbf{Method} & \textbf{Stable (\%)} & \textbf{Slipped (\%)} & \textbf{Fell (\%)} \\
                \midrule
                \multirow{2}{*}{0g}
                    & w/o Current & \textbf{100.0} & 0.0 & 0.0 \\
                    & w/ Current  & \textbf{100.0} & 0.0 & 0.0 \\
                \midrule
                \multirow{2}{*}{150g}
                    & w/o Current & 58.3 & 41.7 & 0.0 \\
                    & w/ Current  & \textbf{83.3} & 16.7 & 0.0 \\
                \midrule
                \multirow{2}{*}{250g}
                    & w/o Current & 16.7 & 58.3 & 25.0 \\
                    & w/ Current  & \textbf{100.0} & 0.0 & 0.0 \\
                \midrule
                \multirow{2}{*}{350g}
                    & w/o Current & 0.0 & 0.0 & 100.0 \\
                    & w/ Current  & \textbf{41.7} & 58.3 & 0.0 \\
                \bottomrule
            \end{tabular}%
        }
        \label{tab:hold-cup-leap}
    \end{minipage}
    \hfill
    \begin{minipage}[t]{0.5\linewidth}
        \centering
        \setlength{\tabcolsep}{3pt}
        \renewcommand{\arraystretch}{0.9}
        \scriptsize
        \caption{Single-card picking results.
    Strict success requires picking exactly one card, while tolerant success allows one or two cards.}
        \vspace{3pt}
        \resizebox{\linewidth}{!}{%
            \begin{tabular}{lcccc}
                \toprule
                \multirow{2}{*}{\textbf{Method}}
                    & \multicolumn{2}{c}{\textbf{Success Rate (\%)}}
                    & \multicolumn{2}{c}{\textbf{Failure Mode (\%)}} \\
                \cmidrule(lr){2-3} \cmidrule(lr){4-5}
                    & \textbf{Strict}
                    & \textbf{Tolerant}
                    & \textbf{2+ Cards}
                    & \textbf{Missed} \\
                \midrule
                w/o Current & 55.8 & 65.4 & 7.7 & 26.9 \\
                w/ Current  & \textbf{76.9} & \textbf{90.4} & \textbf{0.0} & \textbf{9.6} \\
                \bottomrule
            \end{tabular}%
        }
        \label{tab:dex3_card_grasping}
    \end{minipage}
    \vspace{-10pt}
\end{table}

\begin{figure}[!htbp]
    \centering
    \vspace{-15pt}
    \includegraphics[width=1.0\linewidth]{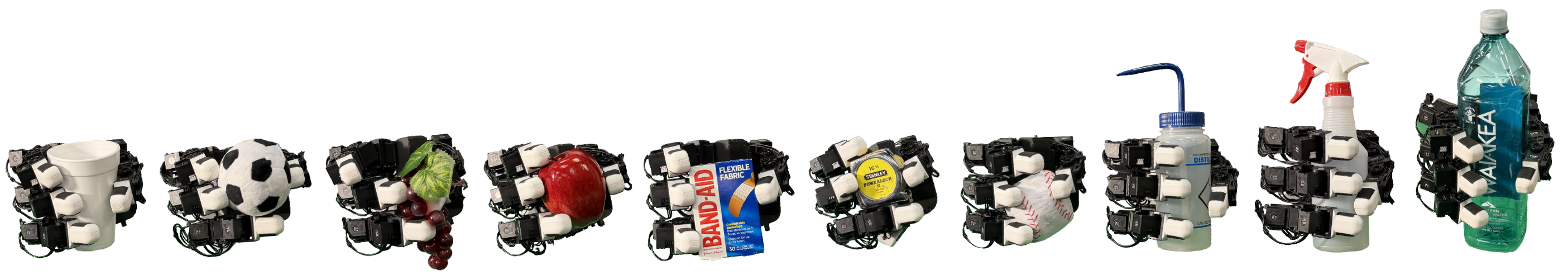}
    \caption{\textbf{Object set for teleoperation training and evaluation.}
    We collect grasping demonstrations on everyday objects with diverse shapes and stiffness, together with free-space hand-motion trajectories, to expose the model to both contact-induced and internal hand-motion currents.}
    \label{fig:object_set}
\end{figure}

The full training objective, with
$\lambda_{\mathrm{cur}}$ balancing the auxiliary current-prediction loss, is:
\begin{equation}
    \mathcal{L}
    =
    \mathcal{L}_{\mathrm{ref}}
    +
    \lambda_{\mathrm{cur}}\mathcal{L}_{\mathrm{cur}} .
\end{equation}

This auxiliary loss regularizes the latent representation to preserve slowly varying current patterns associated with contact and load, while the policy still uses raw current at inference. The auxiliary branch is training-only and adds no sensing or filtering delay at deployment.
\section{Experiments}

We evaluate four contact-rich tasks: teleoperated foam-cup stacking and whiteboard wiping, and policy-learning single-card picking and dynamic bottle holding. Teleoperation compares direct retargeting, a current-free model, and our current-conditioned CRP predictor to test whether motor current is necessary for contact-conditioned reference prediction.

\subsection{Experimental Setup}

Experiments use the LEAP Hand on Franka Research 3 and Dex3 on Unitree G1. Robot state includes hand joints and arm end-effector pose. For teleoperation, \textit{Retargeting} executes the retargeted operator command, \textit{w/o Current} uses the same CRP predictor without current, and \textit{w/ Current} is our full raw-current model. Demonstrations come from operators different from evaluation users, testing transfer rather than memorization of one correction style. For policy learning, the baseline is ACT-style behavior cloning with robot state and object pose; our method adds motor current. During execution, we average the first two predicted CRPs to reduce jitter while preserving responsiveness.

\subsection{Teleoperation Tasks}

\textbf{Foam-cup stacking.}
The operator controls the LEAP Hand and Franka arm to stack four foam cups. We train on 550 trajectories: 50 grasping demonstrations for each of 10 objects with different stiffness, plus 50 free-space hand-motion trajectories. One novice user and one user with 10 minutes of practice each perform 15 trials.

Table~\ref{tab:foam_cup} shows that our method achieves 100.0\% success for both users and reduces completion time. The w/o Current model does not consistently improve over direct teleoperation because sustained wiping requires contact force regulation, not just an average motion prior. Current feedback indicates whether the eraser remains loaded against the board, allowing the CRP to maintain contact with less correction.

\textbf{Board-wiping.}
The operator controls G1 with Dex3 to press an eraser against an inclined whiteboard and wipe it clean. We train on 100 demonstrations and evaluate the same two users over 15 trials each.

As shown in Table~\ref{tab:whiteboard_wiping_results}, our method achieves 100.0\% success for both users and reduces completion time. The w/o Current model does not consistently improve over direct teleoperation, especially for the practiced user, because sustained wiping requires maintaining contact force against changing surface interaction rather than merely following an average motion prior. Current feedback helps the model detect whether the eraser remains properly loaded against the board, allowing the CRP to maintain contact with less manual correction.

\begin{figure}[t]
    \centering
    \includegraphics[width=1.0\linewidth]{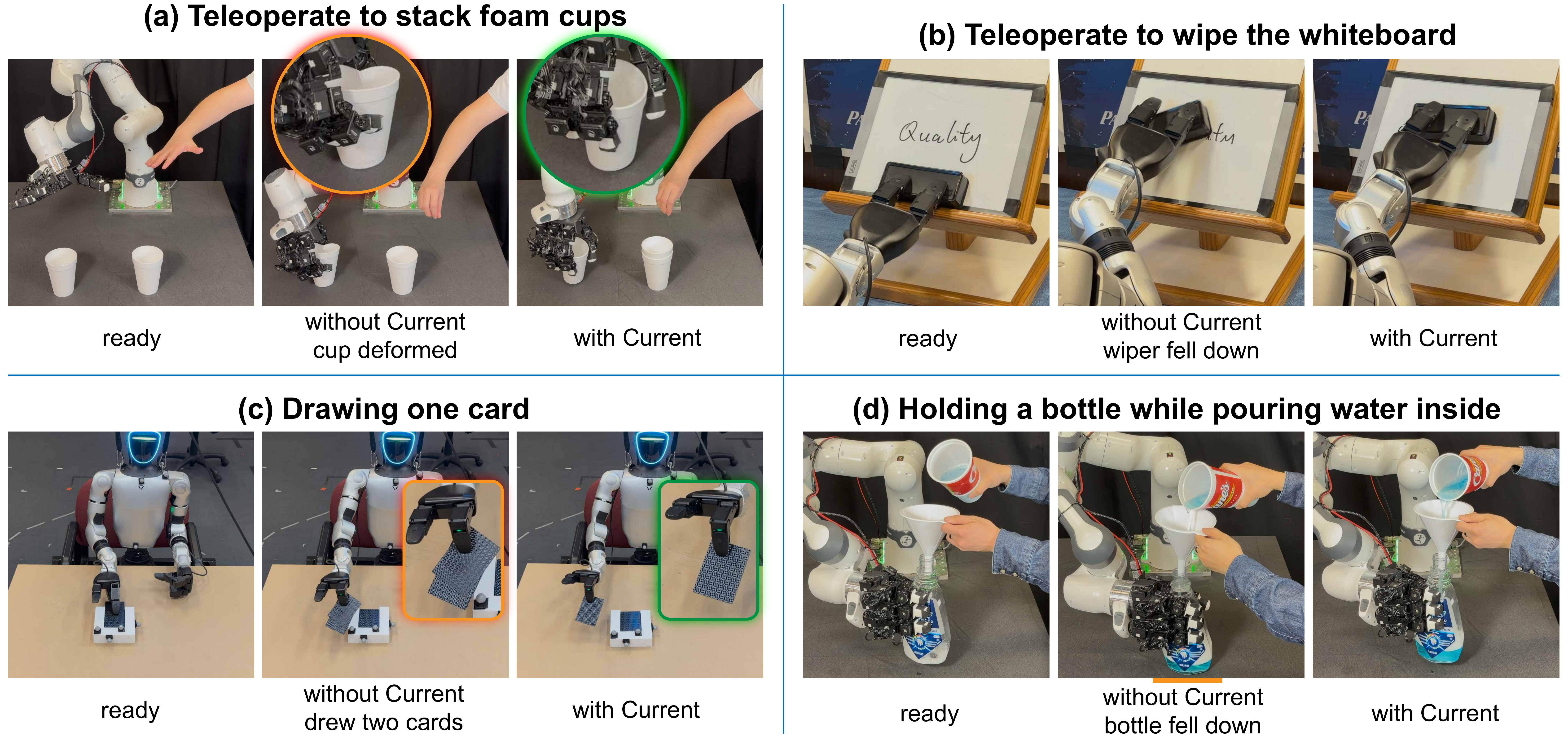}
    \vspace{-15pt}
    \caption{
    \textbf{Qualitative results across teleoperation and policy tasks.}
    Without current-conditioned CRP prediction, direct retargeting or behavior cloning can deform cups, lose wiping contact, draw multiple cards, or drop a dynamically loaded bottle. Motor-current feedback predicts contact-aware CRPs for fragile object handling, sustained surface contact, thin-object retrieval, and dynamic load adaptation.
    }
    \vspace{-6pt}
    \label{fig:experiment_visualization}
\end{figure}

\subsection{Policy Learning Tasks}

\textbf{Dynamic bottle holding.}
A Franka-mounted LEAP Hand grasps a bottle while water is poured into it. We train on 100 demonstrations with approximately 250g poured at a near-constant rate. At test time, water is poured at random rates and evaluated at 0g, 150g, 250g, and 350g, where 350g is outside the training distribution. We conduct 12 trials per load.

This task highlights a subtle advantage of proprioceptive current feedback. Finger flexion currents mainly reflect grip closure rather than vertical load, but the LEAP Hand includes abduction/adduction motors at each finger base. During lateral grasping, increasing bottle weight raises the resistance and current on these joints, providing an implicit load signal.

Table~\ref{tab:hold-cup-leap} shows that both methods are stable at 0g, where little load adaptation is required. As the bottle becomes heavier, the baseline rapidly fails, whereas our method achieves 100\% stability at 250g. Under the OOD 350g load, our method still prevents all drops, although some trials slip. These results suggest that motor current provides a load-sensitive signal, enabling partial generalization beyond the training load and avoiding catastrophic failure.

\textbf{Single-card picking.}
G1 with Dex3 must pick exactly one card from a deck. We train on 150 demonstrations with card SE(3) from VICON during training and inference, and evaluate 52 trials.

Table~\ref{tab:dex3_card_grasping} shows that motor current improves strict success from 55.8\% to 76.9\% and tolerant success from 65.4\% to 90.4\%, while drawing-more-than-two-card failures drop from 7.7\% to 0.0\%. Current helps distinguish sufficient contact from excessive normal force. Without it, the policy infers contact only from position and object pose; with it, the policy receives proprioceptive evidence of contact formation and regulates the CRP more precisely.

\section{Conclusion}

We present a proprioception-driven compliance framework using motor current and joint states. Instead of external tactile/force sensors or explicit wrench estimation, our method predicts a compliance reference position (CRP): a joint-position reference whose PD error generates compliant grasping force. Human-in-the-loop teleoperation provides noisy but valid CRP supervision, while motor current supplies contact-dependent feedback for assisted teleoperation and policy learning. Across foam-cup stacking, whiteboard wiping, single-card picking, and dynamic bottle holding, experiments show safer teleoperation, better contact regulation, and stronger policy robustness.

\noindent \textbf{Limitations.}
Our method relies on motor current being informative about contact, which depends on hand hardware, transmission design, and current measurements. Dynamic load sensing, for example, benefits from the LEAP Hand's finger abduction/adduction motors; hands without such joints may provide weaker shear or vertical-load signals. Demonstrations provide empirical CRP supervision rather than analytically optimal force labels, so performance may depend on demonstration quality and task coverage. Our policy experiments also use task-level object pose when needed and do not address perception or localization; integrating robust perception remains future work.

Our teleoperation experiments include a current-free CRP ablation, showing that kinematics and user intent alone are insufficient for reliable contact-conditioned references. We do not exhaustively ablate every component, such as the auxiliary current loss, execution smoothing, or heuristic current-threshold controllers; more detailed component studies remain future work.

\section*{Acknowledgments}
We gratefully acknowledge SRI International for sponsoring the 6-axis force/torque sensor used in the empirical motor-current/contact-force measurements in this work.

\appendix
\section*{\LARGE Appendix}

\section{Implementation Details}
\label{app:implementation}

This appendix provides additional implementation details for the proposed
current-conditioned compliance reference position (CRP) framework. The system
uses proprioceptive histories, including joint positions and motor current, to
predict position references that are executed by the standard low-level PD
controller. Unless otherwise stated, the real-robot experiments use an
observation horizon of 10 frames and predict a 10-frame future action chunk.

\subsection{Dataset Details}
\label{app:dataset_details}

Table~\ref{tab:app_dataset_stats} summarizes the datasets used in the four
real-robot tasks. For all tasks, trajectories are split into 80\% training and
20\% evaluation sets.

\begin{table}[h]
    \centering
    \caption{Dataset statistics for teleoperation and policy-learning tasks.}
    \label{tab:app_dataset_stats}
    \begin{tabular}{llll}
        \toprule
        Task & Hardware & Demonstrations & Average length / rate \\
        \midrule
        Teleoperated object grasping
        & LEAP Hand + Franka
        & 550
        & 10.2 s at 30.0 Hz \\
        Teleoperated whiteboard wiping
        & Dex3 + Unitree G1
        & 100
        & 8.8 s at 28.6 Hz \\
        Dynamic bottle holding
        & LEAP Hand + Franka
        & 100
        & 14.9 s at 30.0 Hz \\
        Single-card picking
        & Dex3 + Unitree G1
        & 150
        & 8.7 s at 27.9 Hz \\
        \bottomrule
    \end{tabular}
\end{table}

For teleoperated object grasping, we collect 50 grasping demonstrations for
each of 10 objects with different stiffness: foam cup, toy football, grape,
apple, band-aid, steel case ruler, toy baseball, water sprayer I, water sprayer
II, and plastic water bottle. We additionally collect 50 free-space hand-motion
demonstrations in the air. These free-space trajectories expose the model to
motor-current patterns caused by internal hand motion rather than object
contact, helping it separate contact-induced current changes from actuation
currents. All object-grasping data are collected with the LEAP Hand mounted on
a Franka arm.

For teleoperated whiteboard wiping, we collect 100 demonstrations on Dex3. For
dynamic bottle holding, we collect 100 demonstrations with the LEAP Hand and
Franka arm; each demonstration includes grasping an empty bottle and pouring
250 g of water into it. For single-card picking, we collect 150 demonstrations
on Dex3 and G1, where each demonstration draws one card from the initialization
pose.

\subsection{Data Preprocessing}
\label{app:preprocess}

\paragraph{Raw signals.}
Each demonstration is stored as an HDF5 trajectory. For the dexterous hand, we
record measured joint positions $q_t$, target joint positions $q^{\mathrm{cmd}}_t$,
and raw motor currents $I_t$. For embodiments with an arm, we also record the
arm joint positions and arm target positions. In the teleoperation setting, the
operator target is converted into an intent velocity,
\begin{equation}
    v^{\mathrm{intent}}_t =
    q^{\mathrm{cmd}}_t - q^{\mathrm{cmd}}_{t-1}.
    \label{eq:app_intent_velocity}
\end{equation}
The first frame is assigned a zero intent velocity. This representation avoids
feeding the target action directly to the model while preserving whether the
operator intends to close, release, or hold the hand.

\paragraph{Current smoothing for labels and analysis.}
The model receives raw current at inference time, so it does not incur filtering
delay. However, we use an offline-smoothed current as an auxiliary supervision
target and for diagnostic plots. For each joint current trace, the preprocessing
first applies a one-dimensional median filter and then a uniform moving average:
\begin{align}
    I^{\mathrm{med}}_{t,j} &=
    \mathrm{MedianFilter}_{k_m}\left(I_{t,j}\right), \\
    \bar I_{t,j} &=
    \frac{1}{k_u}\sum_{s \in \mathcal{W}_{k_u}(t)}
    I^{\mathrm{med}}_{s,j}.
    \label{eq:app_current_smoothing}
\end{align}
The default implementation uses $k_m=5$ and $k_u=5$ with nearest-boundary
padding. The median filter suppresses isolated communication or PWM spikes,
while the moving average preserves the slower load-dependent trend used by the
auxiliary current loss.

\paragraph{Chunk construction.}
For each valid time index $t$, we construct an observation window
$o_{t-H+1:t}$ of length $H=10$ and an action chunk
$a_{t:t+K-1}$ of length $K=10$. At the beginning of a trajectory, missing
history frames are padded by repeating the first available frame. The
teleoperation observation is
\begin{equation}
    o^{\mathrm{teleop}}_t =
    \left[
    v^{\mathrm{intent}}_t,\;
    q^{\mathrm{robot}}_t,\;
    I_t
    \right],
    \label{eq:app_teleop_obs}
\end{equation}
where $q^{\mathrm{robot}}_t$ concatenates the current hand and arm joint
positions when an arm is used. For autonomous policy learning, the online user
intent term is removed and replaced by task-level perception. In the
single-card picking experiment, we use a VICON motion-capture system to measure
the pose of the card deck relative to the robot base. In the dynamic
bottle-holding experiment, the cup used for pouring is placed at a fixed
position, and the corresponding precomputed $SE(3)$ pose is provided to the
policy. The policy observation can therefore be written as
\begin{equation}
    o^{\mathrm{policy}}_t =
    \left[
    q^{\mathrm{robot}}_t,\;
    I_t,\;
    g_t
    \right].
    \label{eq:app_policy_obs}
\end{equation}
where $g_t$ denotes the task-level object, goal, or pouring pose. For the
Dex3/G1 single-card policy preprocessing, the arm/object pose is represented as
position plus a rotation representation, with the implementation using a 6D
rotation representation by default.

\paragraph{Action space.}
The action label is the demonstrated target position after current-conditioned
regularization by the human operator,
\begin{equation}
    a_t =
    q^{\mathrm{cmd}}_t,
    \label{eq:app_action}
\end{equation}
including hand-joint references and, depending on the embodiment, either arm
joint references or arm/end-effector pose references. This makes the learned
output directly compatible with the position-control interface used by the
corresponding hand-arm system. We evaluated both end-effector pose in $SE(3)$
and arm joint positions as arm-state inputs. Arm joint positions produced
smoother and better tracked teleoperation actions, while policy-learning tasks
showed no consistent performance difference between these two state choices.

\paragraph{Observation and action horizons.}
We tested observation horizons of 8, 10, 12, and 16 frames and did not observe
a significant performance difference. We therefore use $H=10$ as the default
history length, which is long enough to capture recent current and motion
trends while keeping inference lightweight. The action horizon is fixed to
$K=10$ frames. Since action chunking can produce multiple candidate commands
for the same execution time, we study the number of recent predictions used in
the exponential execution average in Section~\ref{app:chunking}.

\paragraph{Small-motion filtering for policy learning.}
Temporal imitation learning can be sensitive to nearly static segments because
the intended future motion becomes ambiguous: many different future references
can be consistent with almost identical observations. For runs in which
small-motion filtering is enabled for policy-learning data, a candidate action
chunk is kept only if its
maximum target displacement exceeds a threshold,
\begin{equation}
    \max_{s \in \{t,\ldots,t+K-1\}}
    \left\|
    q^{\mathrm{cmd}}_s - q^{\mathrm{cmd}}_t
    \right\|_2
    > \epsilon_{\mathrm{move}}.
    \label{eq:app_motion_filter}
\end{equation}
This removes idle clips and clips in which the operator is only holding a
nearly constant command. We apply this filtering to the policy datasets, such
as the single-card picking data, rather than to the teleoperation-assistance
datasets. The appendix figure visualizes one trajectory and highlights which
low-motion intervals are removed by this criterion.

\paragraph{Normalization.}
Observation, action, grasp-intention, and auxiliary current-label tensors are
min-max normalized using statistics computed from the training split:
\begin{equation}
    \tilde x = 2 \frac{x - x_{\min}}{x_{\max}-x_{\min}} - 1.
    \label{eq:app_minmax_norm}
\end{equation}
Dimensions with range smaller than $10^{-6}$ are assigned unit range to avoid
division by zero. The same normalization statistics are saved with each
checkpoint and used during evaluation.

\subsection{Model Architecture}
\label{app:model}

The CRP predictor follows an ACT-style sequence model with a proprioceptive
observation encoder, an action-style encoder used only during training, and a
Transformer decoder over the future action chunk.

\paragraph{Observation encoder.}
The observation history is first transposed into channel-first form and passed
through a temporal convolutional encoder:
\begin{equation}
    z_t = E_\theta(o_{t-H+1:t}).
\end{equation}
The implementation uses three one-dimensional convolution blocks with ReLU
activations, temporal pooling, and a final projection with layer normalization.
The default feature dimension is 64.

\paragraph{ACT-style action encoder and latent style.}
During training, the future action chunk is encoded together with the
observation feature to produce a Gaussian latent style:
\begin{equation}
    (\mu_t, \log \sigma_t^2) = A_\psi(a_{t:t+K-1}, z_t).
\end{equation}
The latent is sampled with the reparameterization trick,
\begin{equation}
    z^{\mathrm{style}}_t =
    \mu_t + \sigma_t \odot \eta,\quad
    \eta \sim \mathcal{N}(0,I).
\end{equation}
At inference time, no future action is available, so the style latent is set to
zero. This preserves the ACT training regularization while keeping deployment
causal.

\paragraph{Action chunk decoder.}
The observation feature and style latent are projected into a model dimension
of 128 and added to learned positional embeddings over the $K=10$ future action
tokens. A four-layer Transformer encoder with four attention heads predicts the
future CRP chunk:
\begin{equation}
    \hat a_{t:t+K-1}
    =
    D_\omega(z_t, z^{\mathrm{style}}_t).
\end{equation}
The default dropout is 0.1.

\paragraph{Auxiliary current head.}
When auxiliary current supervision is enabled, a lightweight MLP predicts the
offline-smoothed current $\bar I_t$ from the observation feature $z_t$. This
branch is used only for training and is removed from the control loop at
deployment, so the deployed policy still conditions on raw current without
phase delay.

\begin{table}[t]
    \centering
    \caption{Default model and training hyperparameters.}
    \label{tab:app_hyperparams}
    \begin{tabular}{ll}
        \toprule
        Hyperparameter & Value \\
        \midrule
        Observation horizon $H$ & 10 frames \\
        Action horizon $K$ & 10 frames \\
        Optimizer & AdamW \\
        Learning rate & $1\times 10^{-4}$ \\
        Batch size & 32 \\
        Epochs & 300 \\
        Observation feature dimension & 64 \\
        Transformer model dimension & 128 \\
        Transformer layers & 4 \\
        Attention heads & 4 \\
        Dropout & 0.1 \\
        KL weight $\lambda_{\mathrm{KL}}$ & $1\times 10^{-5}$ \\
        Auxiliary current weight $\lambda_{\mathrm{cur}}$ & 0.1 \\
        KL annealing length & 100 epochs \\
        Checkpoint interval & 2 epochs \\
        Evaluation interval & 1 epoch \\
        \bottomrule
    \end{tabular}
\end{table}

\subsection{Training Objective and Schedule}
\label{app:loss_schedule}

The main action loss is the mean-squared error between the predicted action
chunk and the demonstrated CRP chunk:
\begin{equation}
    \mathcal{L}_{\mathrm{ref}}
    =
    \left\|
    \hat a_{t:t+K-1} - a_{t:t+K-1}
    \right\|_2^2.
    \label{eq:app_ref_loss}
\end{equation}
The ACT-style latent is regularized with the KL divergence between the encoded
posterior and a unit Gaussian prior:
\begin{equation}
    \mathcal{L}_{\mathrm{KL}}
    =
    -\frac{1}{2}
    \mathbb{E}
    \left[
    \sum_d
    \left(
    1+\log\sigma_{t,d}^{2}
    -\mu_{t,d}^{2}
    -\sigma_{t,d}^{2}
    \right)
    \right].
    \label{eq:app_kl}
\end{equation}
If auxiliary current prediction is enabled, the current head is trained with
\begin{equation}
    \mathcal{L}_{\mathrm{cur}}
    =
    \left\|
    \hat{\bar I}_t - \bar I_t
    \right\|_2^2.
    \label{eq:app_current_loss}
\end{equation}
The total loss at epoch $e$ is
\begin{equation}
    \mathcal{L}
    =
    \mathcal{L}_{\mathrm{ref}}
    +
    \lambda_{\mathrm{KL}}(e)\mathcal{L}_{\mathrm{KL}}
    +
    \lambda_{\mathrm{cur}}\mathcal{L}_{\mathrm{cur}}.
    \label{eq:app_total_loss}
\end{equation}
The KL weight is linearly annealed:
\begin{equation}
    \lambda_{\mathrm{KL}}(e)
    =
    \lambda_{\mathrm{KL}}^{\max}
    \min\left(1,\frac{e+1}{E_{\mathrm{anneal}}}\right),
    \label{eq:app_kl_anneal}
\end{equation}
where the default $\lambda_{\mathrm{KL}}^{\max}=10^{-5}$ and
$E_{\mathrm{anneal}}=100$ epochs. This schedule prevents the latent from being
over-regularized early in training while still discouraging uncontrolled action
style variation later.

\subsection{Action Chunking and Execution Aggregation}
\label{app:chunking}

The model predicts a 10-frame sequence at each control step. During execution,
multiple recent predictions may provide candidate CRPs for the current time
index. We aggregate the most recent $M$ candidates with an exponential
average:
\begin{equation}
    q^{\mathrm{exec}}_t =
    \frac{
    \sum_{m=0}^{M-1} \alpha^m \hat q^{(t-m)}_{t}
    }{
    \sum_{m=0}^{M-1} \alpha^m
    },
    \label{eq:app_temporal_agg}
\end{equation}
where $\hat q^{(t-m)}_{t}$ denotes the CRP for time $t$ predicted $m$ control
steps earlier. We use a short aggregation window so that the command is smooth
enough for hardware execution but still responsive to sudden current changes
caused by contact.

\begin{table}[t]
    \centering
    \caption{Execution aggregation ablation for dynamic bottle holding. Each
    setting is evaluated over 10 trials. Grasp success measures whether the
    robot first establishes a stable grasp on the empty bottle. Hold success
    measures whether the robot keeps the bottle from 0 g to 250 g of poured
    water after a successful grasp.}
    \label{tab:app_exec_aggregation}
    \begin{tabular}{ccc}
        \toprule
        Aggregated frames $M$ & Grasp success & Hold success, 0 g to 250 g \\
        \midrule
        1 & 40\% & 90\% \\
        2 & 70\% & 100\% \\
        3 & 90\% & 100\% \\
        4 & 100\% & 70\% \\
        \bottomrule
    \end{tabular}
\end{table}

Table~\ref{tab:app_exec_aggregation} shows the effect of the execution
aggregation window. With $M=1$, the command uses only the newest prediction.
This makes the controller highly responsive, but the hand can enter small
oscillations during grasp acquisition, reducing initial grasp success to
40\%. Once a stable grasp is established, however, the current-conditioned CRP
can still maintain the bottle in 90\% of the pouring trials. Increasing the
window to $M=2$ improves grasp acquisition to 70\% by damping frame-to-frame
command jitter while preserving enough responsiveness to handle the increasing
load. Using $M=3$ further improves grasp success to 90\% and still maintains
100\% hold success from 0 g to 250 g. With $M=4$, the initial grasp becomes
very smooth and reaches 100\% success, but the longer averaging window delays
the response to dynamic load changes, reducing hold success to 70\%. We
therefore use a short aggregation window, with $M=3$ providing the best balance
between stable grasp acquisition and load-adaptive holding in these trials.

\section{KL Weight Ablations}
\label{app:kl_ablation}

We ablate the KL weight used by the ACT-style action encoder for teleoperation
and bottle grasping. Each setting is evaluated over 10 real-robot trials.

\begin{table}[t]
    \centering
    \caption{Teleoperation fist-closing success under different KL weights.
    Each setting is evaluated over 10 trials.}
    \label{tab:app_kl_fist}
    \begin{tabular}{ccc}
        \toprule
        $\lambda_{\mathrm{KL}}^{\max}$ & Successful trials / 10 & Success rate \\
        \midrule
        $0$ & 10/10 & 100\% \\
        $1\times 10^{-6}$ & 8/10 & 80\% \\
        $1\times 10^{-5}$ & 6/10 & 60\% \\
        $1\times 10^{-4}$ & 3/10 & 30\% \\
        \bottomrule
    \end{tabular}
\end{table}

\begin{table}[t]
    \centering
    \caption{Bottle-grasping success under different KL weights. Each setting
    is evaluated over 10 trials.}
    \label{tab:app_kl_cup}
    \begin{tabular}{ccc}
        \toprule
        $\lambda_{\mathrm{KL}}^{\max}$ & Successful trials / 10 & Success rate \\
        \midrule
        $0$ & 9/10 & 90\% \\
        $1\times 10^{-6}$ & 10/10 & 100\% \\
        $1\times 10^{-5}$ & 10/10 & 100\% \\
        $1\times 10^{-4}$ & 9/10 & 90\% \\
        \bottomrule
    \end{tabular}
\end{table}

The teleoperation result shows a clear sensitivity to the KL weight. With
$\lambda_{\mathrm{KL}}^{\max}=0$, the model follows the user's closing intent
most directly and succeeds in all 10 trials. As the KL weight increases, the
latent action style is pulled more strongly toward the prior, making the
predicted CRPs more conservative and less able to follow the user's online
motion. This reduces fist-closing success from 100\% to 30\% as
$\lambda_{\mathrm{KL}}^{\max}$ increases from 0 to $10^{-4}$.

In contrast, bottle-grasping success is much less sensitive to the tested KL
weights. All settings achieve either 90\% or 100\% success, suggesting that
this policy task is dominated by the proprioceptive and task-state inputs
rather than fine-grained user-following behavior. The default
$\lambda_{\mathrm{KL}}^{\max}=10^{-5}$ therefore provides stable bottle
grasping while retaining the regularization benefits of the ACT-style latent
for sequence prediction.

\section{Motor Current, Joint Torque, and Contact Force}
\label{app:current_torque_force}

Motor current is related to actuator torque, but the measured current is not a
pure contact-force signal. This section makes explicit the main physical terms
that appear in the measurement and explains why we learn a CRP directly rather
than analytically converting current into contact force.

\subsection{Current-to-Torque Model}
\label{app:current_to_torque}

For a motorized joint $j$, the electromagnetic torque can be approximated by
\begin{equation}
    \tau^{\mathrm{em}}_{j,t}
    =
    k_{\tau,j}
    \left(
    I_{j,t} - I^{0}_{j}
    \right),
    \label{eq:app_motor_torque}
\end{equation}
where $k_{\tau,j}$ is the motor torque constant and $I^{0}_{j}$ is a current
offset caused by electronics, calibration bias, and static preload. The torque
available at the joint is affected by transmission and internal losses:
\begin{equation}
    \tau^{\mathrm{joint}}_{j,t}
    =
    \eta_j r_j \tau^{\mathrm{em}}_{j,t}
    -
    \tau^{\mathrm{fric}}_{j,t}
    -
    \tau^{\mathrm{dyn}}_{j,t}
    -
    \tau^{\mathrm{bias}}_{j,t}
    +
    \epsilon^{\tau}_{j,t}.
    \label{eq:app_joint_torque}
\end{equation}
Here $r_j$ is the transmission ratio, $\eta_j$ is transmission efficiency,
$\tau^{\mathrm{fric}}$ includes Coulomb and viscous friction,
$\tau^{\mathrm{dyn}}$ includes inertial, Coriolis, and motor back-EMF effects,
$\tau^{\mathrm{bias}}$ captures gravity, cable tension, tendon preload, gear
backlash, and unmodeled elastic elements, and $\epsilon^\tau$ is stochastic
measurement and actuation noise.

A commonly used decomposition is
\begin{equation}
    \tau^{\mathrm{fric}}_{j,t}
    =
    b_j \dot q_{j,t}
    +
    c_j\,\mathrm{sgn}(\dot q_{j,t})
    +
    \tau^{\mathrm{strib}}_{j,t},
    \label{eq:app_friction}
\end{equation}
where $b_j$ is viscous friction, $c_j$ is Coulomb friction, and
$\tau^{\mathrm{strib}}$ captures low-velocity Stribeck effects. Backlash and
tendon compliance can be represented as history-dependent terms:
\begin{equation}
    \tau^{\mathrm{bias}}_{j,t}
    =
    g_j(q_t)
    +
    \tau^{\mathrm{cable}}_j(q_t, h_t)
    +
    \tau^{\mathrm{backlash}}_j(h_t),
    \label{eq:app_bias}
\end{equation}
where $h_t$ denotes recent motion history. These terms are difficult to identify
accurately for low-cost dexterous hands and may drift with temperature, wear,
and cable routing.

\subsection{From Joint Torque to Contact Force}
\label{app:torque_to_contact}

If the hand kinematics and contact location were exactly known, contact wrench
$f_t$ could be related to joint torque through the contact Jacobian:
\begin{equation}
    \tau^{\mathrm{contact}}_t
    =
    J_c(q_t)^\top f_t.
    \label{eq:app_contact_jacobian}
\end{equation}
In practice, this inverse problem is ill-conditioned for dexterous hands. The
contact location may be unknown, multiple fingers may contact the object
simultaneously, object compliance changes the local normal direction, and
sliding or rolling changes the effective Jacobian. Therefore, the measured
current contains a mixture of useful contact information and nuisance effects:
\begin{equation}
    I_t
    =
    \Phi_{\mathrm{contact}}(q_t,\dot q_t,f_t)
    +
    \Phi_{\mathrm{internal}}(q_t,\dot q_t,h_t)
    +
    \epsilon^I_t.
    \label{eq:app_current_decomposition}
\end{equation}
The first term is the contact-dependent part we want to exploit. The second
term includes internal friction, motor dynamics, cable effects, backlash,
gravity, and thermal drift. The final term includes random sensor noise,
quantization, communication spikes, and unmodeled disturbances.

This decomposition also clarifies what can and cannot be learned from data.
Many hardware- and manufacturing-dependent effects are not analytically known
but are repeatable for a fixed robot, such as joint-dependent current offsets,
transmission efficiency, typical cable friction, or backlash patterns under
similar motion histories. A sufficiently diverse dataset can let the model
absorb these systematic terms into the learned mapping from
$(q_{t-H+1:t}, I_{t-H+1:t})$ to CRPs. In contrast, truly random terms, such as
isolated communication spikes, quantization noise, or non-repeatable impacts,
cannot be predicted deterministically from the observation history. They can
only be attenuated statistically through temporal context, smoothing used for
auxiliary labels, robust training data, and conservative execution averaging.
We include this distinction because it explains both the strength and the
limitation of current-based tactile feedback: current provides useful
contact-correlated information, but it is not a noise-free tactile force sensor.

\subsection{Why Learn CRPs Instead of Explicit Force}
\label{app:why_learn_crp}

Our method does not require a calibrated conversion from current to force.
Instead, it learns a mapping
\begin{equation}
    \hat q^{c}_{\mathrm{ref},t}
    =
    f_\theta(q_{t-H+1:t}, I_{t-H+1:t}, u_t),
    \label{eq:app_learned_crp}
\end{equation}
where $u_t$ denotes user intent in teleoperation or task-level perception in
policy learning. The learned mapping can use the repeatable part of the current
signal, such as increased motor load during contact, while learning to ignore
or average over nuisance terms that are not predictive of successful CRPs. The
PD controller then converts the predicted position reference into interaction
torque:
\begin{equation}
    \tau_t =
    K_p
    \left(
    \hat q^{c}_{\mathrm{ref},t} - q_t
    \right)
    -
    K_d \dot q_t.
    \label{eq:app_pd_again}
\end{equation}
This position-reference formulation matches the command interface of common
dexterous hands and avoids relying on an explicit wrench estimator, contact
Jacobian inversion, or torque-control interface.

\bibliography{main}

\end{document}